\documentclass[11pt]{article}

\usepackage[final,nohyperref]{acl}

\usepackage{latexsym}
\usepackage[T2A,T1]{fontenc}
\usepackage[utf8]{inputenc}
\usepackage{times}
\usepackage{inconsolata}

\DeclareUnicodeCharacter{04E2}{\={\CYRI}}
\DeclareUnicodeCharacter{04E3}{\={\cyri}}
\DeclareUnicodeCharacter{04EE}{\={\CYRU}}
\DeclareUnicodeCharacter{04EF}{\={\cyru}}
\usepackage{microtype}
\usepackage{graphicx}
\usepackage{booktabs}
\usepackage{url}
\usepackage{enumitem}
\setlist{nosep,leftmargin=*}

\newcommand{\SixStockBelTop}{95}
\newcommand{\SixStockBelTopL}{A}

\newcommand{\SixLightBelTop}{48}

\newcommand{\SixLightEnMTop}{45}
\newcommand{\SixLightEnMTopL}{C}

\newcommand{\SixBaseEnM}{48.8}

\newcommand{\SixFinalTl}{1.95}
\newcommand{\SixFinalBelTop}{55}

\newcommand{\SixFinalBel}{30.9}

\newcommand{\EightStockBel}{57.9}

\newcommand{\EightStockEnM}{73.5}

\newcommand{\EightBooksTl}{1.83}

\newcommand{\EightCptTl}{1.58}

\newcommand{\EightSftBel}{69.3}

\newcommand{\EightTaxBelTop}{40}

\newcommand{\SixLightDTlAbs}{1.45}
\newcommand{\SixLightDBel}{+4.2}
\newcommand{\SixLightDBelAbs}{4.2}
\newcommand{\SixLightDBelLo}{+0.9}

\newcommand{\SixLightDBelHi}{+7.8}

\newcommand{\SixLightDEnMAbs}{9.4}

\newcommand{\OneSevVOneDTlAbs}{1.29}
\newcommand{\OneSevVOneDBel}{$-$0.8}

\newcommand{\OneSevVOneDEnMAbs}{3.3}

\newcommand{\OneSevVTwoDBel}{+0.0}

\newcommand{\OneSevVTwoDEnMAbs}{3.7}

\newcommand{\FourLightDTlAbs}{0.86}

\newcommand{\FourLightDBelAbs}{3.7}
\newcommand{\FourLightDBelLo}{$-$6.9}

\newcommand{\FourLightDBelHi}{$-$0.2}

\newcommand{\SixSftADEnMAbs}{6.2}

\newcommand{\SixCptDEnMAbs}{8.5}

\newcommand{\SixFinalDEnMAbs}{7.6}

\newcommand{\EightLightDTlAbs}{0.60}
\newcommand{\EightLightDBel}{+0.1}

\newcommand{\EightLightDBelLo}{$-$3.3}

\newcommand{\EightLightDBelHi}{+3.2}

\newcommand{\EightBooksDBelAbs}{7.4}
\newcommand{\EightBooksDBelLo}{+4.2}

\newcommand{\EightBooksDBelHi}{+10.7}

\newcommand{\EightCptDBelAbs}{9.9}
\newcommand{\EightCptDBelLo}{+6.3}

\newcommand{\EightCptDBelHi}{+13.4}

\newcommand{\EightSftDBelAbs}{11.4}

\newcommand{\EightSftDEnM}{$-$1.2}

\newcommand{\EightSftDEnMLo}{$-$3.3}
\newcommand{\EightSftDEnMLoAbs}{3.3}
\newcommand{\EightSftDEnMHi}{+0.9}

\newcommand{\NBel}{900}
\newcommand{\NEnM}{1,003}
\newcommand{\NMmluSubset}{1,003}
\newcommand{\NMmluFull}{14,042}
\newcommand{\MmluStride}{14}
\newcommand{\NPassages}{488}
\newcommand{\NPassageTokens}{139,477}
\newcommand{\NModels}{21}

\newcommand{\NStock}{5}
\newcommand{\NTrainedDistinct}{15}
\newcommand{\BelGoldTop}{28}

\newcommand{\EightMathVsSftBel}{+0.2}

\newcommand{\EightMathVsSftEnMAbs}{2.7}
\newcommand{\EightMultiVsSftBel}{+0.8}

\newcommand{\EightMultiVsSftEnM}{$-$1.5}

\newcommand{\EightTaxVsSftBelAbs}{3.0}

\newcommand{\EightTaxVsSftEnMAbs}{0.9}

\newcommand{\MathMergedSame}{identical}
\newcommand{\NADocs}{41,877}
\newcommand{\NANews}{26,601}
\newcommand{\NABooks}{15,276}
\newcommand{\ACutSmall}{512}
\newcommand{\ACutSix}{1,024}
\newcommand{\ATokensM}{21.4}
\newcommand{\NCptChunks}{183,050}
\newcommand{\CptChunkLen}{2,048}
\newcommand{\NCptBooks}{39,100}
\newcommand{\NCptNews}{128,463}
\newcommand{\NCptWiki}{15,487}
\newcommand{\CptEpochs}{4}
\newcommand{\CptBooksEpochs}{3}
\newcommand{\NFullFt}{3}
\newcommand{\NThm}{1,227,427}
\newcommand{\NThmFiles}{11}
\newcommand{\NThmReason}{6,600}
\newcommand{\ThmReasonPct}{0.5}
\newcommand{\ThmEpochs}{3}
\newcommand{\NThmEval}{698}
\newcommand{\NMaths}{17,446}
\newcommand{\NExam}{10,982}
\newcommand{\NTjkData}{8,748}
\newcommand{\NTax}{70,186}
\newcommand{\TaxTokensM}{45.8}
\newcommand{\NTaxHeldDeleted}{3,510}
\newcommand{\NSixSft}{1,122,175}
\newcommand{\NSixSources}{12}
\newcommand{\SixCptEpochs}{2}
\newcommand{\SixSftOneEpochs}{3}
\newcommand{\SixSftTwoEpochs}{5}
\newcommand{\SixSftThreeEpochs}{2}
\newcommand{\SinqTensors}{147}
\newcommand{\SinqGb}{2.49}

\newcommand{\ThmChatPct}{99.5}
\newcommand{\CptTokensM}{375}
\newcommand{\CptOverA}{18}
\newcommand{\BooksLossFirst}{2.73}
\newcommand{\BooksLossLast}{1.73}
\newcommand{\BooksSteps}{3,650}
\newcommand{\BwnLossFirst}{0.96}
\newcommand{\BwnLossLast}{0.95}
\newcommand{\SixLightRank}{16}

\newcommand{\FourLightRank}{8}

\newcommand{\SixFinalRank}{64}
\newcommand{\EightLightRank}{8}
\newcommand{\EightBooksRank}{32}
\newcommand{\EightCptRank}{16}

\newcommand{\ESft}{e3}
\newcommand{\EMath}{e3.1}
\newcommand{\RepoSft}{\texttt{SaidzodaEng/\allowbreak Ameena\_Qwen3-8B\_e3}}
\newcommand{\RepoSixFinal}{\texttt{SaidzodaEng/\allowbreak Ameena\_Qwen3-0.6B\_LoRA\allowbreak \_e8\_p2\_e3}}
\newcommand{\RepoResults}{\url{https://huggingface.co/datasets/Tohirju/Saidzoda-Qwen3-Tajik-Evals}}

\title{Adapting Qwen3 to Tajik from 0.6B to 8B:\\Lower Text Loss Did Not Mean Better Understanding}

\author{Tohir Saidzoda \and Madina Abdusalomzoda \and Nasimjon Rajabov \\
  \textbf{Akhrorkhon Ikromov} \and \textbf{Komron Giyosiev} \and \textbf{Abubakr Abdullozoda} \\
  \textbf{Amin Jalilov} \and \textbf{Jonibek Kishvarov} \and \textbf{Rahmonali Sharipov} \and \textbf{Mariyamkhon Shodyeva} \\
  Saidzoda AI Lab, Saidzoda Engineering Company, Dushanbe, Tajikistan \\
  \texttt{tohir@saidzoda.com}}

\begin{document}
\maketitle

\begin{abstract}
Small LoRA adapters trained on Tajik news and books lowered Qwen3's Tajik text loss at every size we tried, from 0.6B to 8B, by \EightLightDTlAbs{} to \SixLightDTlAbs{} nats per token, but none of them improved Tajik reading comprehension where it can be measured.
On Belebele the 1.7B and 8B adapters left accuracy unchanged and the 4B adapter lowered it by \FourLightDBelAbs{} points; at 0.6B the score rose by \SixLightDBelAbs{} points but sits near chance and coincides with a change in which answer letter the model prefers, so we do not read it as understanding.
At 8B, continued pretraining on about \CptTokensM{} million tokens of raw Tajik text was followed by a Belebele gain of \EightCptDBelAbs{} points.
After a further supervised fine-tuning stage, our main model, \ESft, is \EightSftDBelAbs{} points above Qwen3-8B on Belebele; its English MMLU changed by \EightSftDEnM{} points, and a loss larger than \EightSftDEnMLoAbs{} points is excluded.
The same heavy recipe applied to Qwen3-0.6B-Base brought its text loss close to that of the 8B model after books-only pretraining, while its multiple-choice answers stayed near chance and its English MMLU fell by \SixFinalDEnMAbs{} points.
We report \NTrainedDistinct{} trained models and their \NStock{} starting models from June to October 2025, scored in 2026 on one harness with paired intervals, and we report what went wrong: lost training records, a thinking-mode dataset with almost no reasoning in it, later branches whose training data was never documented, and a 4-bit release that contains no transformer layers.

\end{abstract}

\section{Introduction}
\label{sec:intro}

Tajik, the state language of Tajikistan, is written in a Cyrillic alphabet with six letters that Russian does not have.
Open language models rarely list it among their supported languages.
In June 2025 we set out to make Qwen3 \citep{yang2025qwen3} usable in Tajik with the resources a small lab has: single GPUs, a few days per run, and Tajik text collected from news sites, books and Wikipedia.

We tried two recipes.
The light recipe, used first, trains a small attention-only LoRA adapter \citep{hu2021lora} on a 4-bit model \citep{dettmers2023qlora} with at most about \ATokensM{} million tokens of news and books per epoch.
The heavy recipe adds continued pretraining (CPT) on about \CptTokensM{} million tokens of raw Tajik text, followed by supervised fine-tuning (SFT) on more than a million Tajik chat examples.
We applied the light recipe at 0.6B, 1.7B, 4B and 8B, and the heavy recipe at 8B and later again at 0.6B.

In 2025 we judged these models by reading their answers.
They all wrote Tajik, so all of them looked like progress.
None was scored on a held-out benchmark at the time.
For this report we scored the \NTrainedDistinct{} trained models and their \NStock{} starting models in 2026 on one harness, each trained model against the model it started from, and the numbers tell a different story from the one we believed.

Our contributions are:
\begin{enumerate}
\item \textbf{A measured record of five months of Tajik adaptation.} Models from 0.6B to 8B, scored on Tajik reading comprehension, English knowledge, and Tajik text loss, with paired intervals against each model's own starting point (Section~\ref{sec:results}).
\item \textbf{Lower text loss is not understanding.} Light adapters lower Tajik text loss at every size without improving reading comprehension at 1.7B, 4B or 8B; the 0.6B scores sit near chance. Only the 8B models with continued pretraining improved on it (Figure~\ref{fig:fluency}). At 0.6B even the heavy recipe lowered text loss without moving the multiple-choice scores away from chance, and we show that the scores of the smallest models are dominated by a preference for one answer letter (Table~\ref{tab:bias}).
\item \textbf{A record of what failed.} Lost epochs, a thinking-mode dataset that was \ThmChatPct\% ordinary chat, a tax test set deleted before it was used, later branches whose training data was never documented, and a quantised release missing its transformer layers (Section~\ref{sec:failures}).
\end{enumerate}

\section{Related Work}
\label{sec:related}

\paragraph{Adapting models to a new language.}
The usual route is continued pretraining on raw text in the new language, often followed by instruction tuning \citep{cui2023chinesellama}.
Continued pretraining shifts the model's distribution and can make it forget what it knew \citep{luo2023forgetting}; learning-rate rewarming and replay reduce the loss \citep{gupta2023rewarm,ibrahim2024simple}.
LoRA \citep{hu2021lora} and its 4-bit form QLoRA \citep{dettmers2023qlora} make this affordable on one GPU.
LoRA learns less from new data than full fine-tuning but also forgets less \citep{biderman2024lora}, which matters when the goal is a new language rather than a new style.
Multilingual pretraining helps low-resource languages up to a point and then competes with them \citep{chang2023curse}.

\paragraph{Tajik.}
Large web corpora now include Tajik \citep{kudugunta2023madlad400,penedo2025fineweb2}, and recent work has released Tajik text-processing tools and compared parameter-efficient fine-tuning methods on Tajik \citep{arabov2026tajiknlp,arabov2026tajikweb}.
Belebele \citep{bandarkar2024belebele}, built on FLORES passages \citep{nllb2022}, covers Tajik reading comprehension.

\paragraph{Evaluating adapted models.}
Studies of language adaptation usually report held-out loss and downstream accuracy together.
Our results add a case where the two disagree strongly: text loss improves in every run, while accuracy, measured by Belebele, improves only when the model is large enough and the training text is large enough.

\paragraph{Quantisation.}
SINQ quantises weights to 4 bits without calibration data by balancing row and column scales \citep{muller2025sinq}.
We used it in 2025 and report in Section~\ref{sec:failures} why our quantised uploads cannot be evaluated.

\section{Training Data}
\label{sec:data}

\paragraph{News and books (light recipe).}
\NANews{} news articles and \NABooks{} book chunks, \NADocs{} documents in total.
Each document was wrapped as one chat turn, with the title as the user message and the text as the answer, and the loss covered the whole text.
This is continued pretraining inside a chat template, not instruction tuning.
Texts were cut to \ACutSmall{} tokens for the 1.7B and 4B runs, which gives at most about \ATokensM{} million tokens per epoch. The data of the 0.6B adapter we evaluate was not recorded; an earlier 0.6B run used news only, cut to \ACutSix{} tokens.

\paragraph{CPT corpus (heavy recipe).}
\NCptChunks{} chunks of \CptChunkLen{} tokens each, about \CptTokensM{} million tokens: \NCptBooks{} from books, \NCptNews{} from news and \NCptWiki{} from Tajik Wikipedia.
The text is raw, cut into fixed windows without document boundaries.
The Wikipedia chunks still contain page navigation text.
One epoch of this corpus is about \CptOverA{} times one epoch of the light-recipe data.

\paragraph{SFT data.}
For 8B: \NThm{} chat examples from \NThmFiles{} files, including question answering, dictionary entries, Wikipedia rewritten as dialogue, sentence tasks, instructions and official documents.
It was called a ``thinking mode'' dataset, but only about \NThmReason{} examples (\ThmReasonPct\%) contain reasoning.
For 0.6B: \NSixSft{} examples from \NSixSources{} sources, which overlap with the 8B set but drop the Wikipedia dialogues and add university entrance-exam questions and questions about Tajikistan.

\paragraph{Branch data.}
Three later 8B branches added: a maths set of \NMaths{} examples whose contents were never documented; \NExam{} university entrance-exam questions with model-written explanations and \NTjkData{} questions about Tajikistan; and \NTax{} examples on the Tajik tax code (\TaxTokensM{} million tokens).

\section{Models and Recipes}
\label{sec:recipes}

All training used Unsloth with LoRA adapters.
Table~\ref{tab:cells} summarises the four groups.

\paragraph{Light recipe, June 2025.}
Qwen3-0.6B, Qwen3-1.7B and Qwen3-4B (the released chat models) were loaded in 4 bits and given attention-only adapters of rank \FourLightRank{} to \SixLightRank{} on the news and books data.
The 1.7B model has two versions: v1 after \NADocs{} documents, and v2 after a further pass over a subset.
An adapter of rank \EightLightRank{} with the same settings as the 4B run was trained on Qwen3-8B; its training data was not recorded.

\paragraph{Heavy recipe at 8B, July to August 2025.}
We first ran CPT on books only, with a rank-\EightBooksRank{} adapter on the 16-bit model for \CptBooksEpochs{} epochs (logged training loss \BooksLossFirst{} at the start and \BooksLossLast{} at step \BooksSteps).
Separately, a rank-\EightCptRank{} adapter on all projections of the 4-bit model was trained for \CptEpochs{} epochs on the full CPT corpus; only the last epoch has a training record, and the middle epochs are known only from saved checkpoints.
\NFullFt{} attempts at full fine-tuning on the books data were stopped within the first steps and nothing was saved.
SFT then continued the same adapter for \ThmEpochs{} epochs on the \NThm{} chat examples, and the result was merged into Qwen3-8B as \ESft{} (\RepoSft).
Three branches continue from there: maths (\EMath), a further adapter on the merged \EMath{} model with the entrance-exam and Tajikistan data (MultiStage), and the tax code.
Because all 8B stages continue one adapter (the maths notebook loads the SFT adapter as its starting point), the \EMath{} adapter applied to Qwen3-8B and the merged \EMath{} file should be the same model; their answers are \MathMergedSame{} on every test.

\paragraph{Heavy recipe at 0.6B, September to October 2025.}
We returned to the smallest size with Qwen3-0.6B-Base and a rank-\SixFinalRank{} adapter on all projections: SFT for \SixSftOneEpochs{} epochs, SFT for about \SixSftTwoEpochs{} more (stopped just before the end), CPT on the same corpus as the 8B model for \SixCptEpochs{} epochs, and a final SFT of \SixSftThreeEpochs{} epochs that started from a checkpoint of an earlier, stopped attempt (\RepoSixFinal).
In the SFT stages the loss covered only the answers.
The mask also excluded the end-of-turn token, so the model never learned to stop, which may explain the endless replies we saw in its samples.

\begin{table*}[t]
\centering\footnotesize\setlength{\tabcolsep}{4pt}
\begin{tabular}{p{1.9cm}p{2.9cm}p{2.2cm}p{3.5cm}p{3.5cm}}
\toprule
 & Light, small (June) & Light, 8B (June) & Heavy, 0.6B (Sep to Oct) & Heavy, 8B (Jul to Aug) \\
\midrule
Start model & Qwen3 0.6B, 1.7B, 4B & Qwen3-8B & Qwen3-0.6B-Base & Qwen3-8B \\
Text & news and books, \NADocs{} documents (0.6B: not recorded) & not recorded & \NSixSft{} chat examples + raw text & raw text, then \NThm{} chat examples \\
Tokens per epoch & about \ATokensM M & not recorded & about \CptTokensM M (CPT stage) & about \CptTokensM M (CPT stage) \\
Order & one adapter run per model & not recorded & SFT, SFT, CPT, SFT & CPT (\CptEpochs{} epochs), then SFT (\ThmEpochs) \\
Loss on & whole text & not recorded & answer only & whole text \\
Adapter & rank \FourLightRank{} to \SixLightRank, attention only & rank \EightLightRank, attention only & rank \SixFinalRank, all projections & rank \EightCptRank, all projections \\
Precision & 4-bit & 4-bit & 16-bit, later 4-bit & 4-bit \\
\bottomrule
\end{tabular}
\caption{The four groups of models compared in this report. The groups differ in more than size and recipe, so the comparison is not controlled. Values come from the training notebooks and adapter files (Section~\ref{sec:recipes}).}
\label{tab:cells}
\end{table*}

\section{Evaluation}
\label{sec:eval}

No 2025 model had a held-out score, so we scored all of them in October 2026 with one script on one NVIDIA A10.

\paragraph{Tests.}
\begin{itemize}
\item \textbf{Belebele Tajik} \citep{bandarkar2024belebele}: \NBel{} reading-comprehension questions about short passages. This is our measure of understanding.
\item \textbf{English MMLU} \citep{hendrycks2021mmlu}: \NEnM{} of the \NMmluFull{} test questions (every \MmluStride th, which keeps all subjects in proportion), to see what the models forgot.
\item \textbf{Tajik text loss}: the mean negative log-likelihood per token over the \NPassages{} distinct Tajik Belebele passages (\NPassageTokens{} tokens). All models share the Qwen3 tokenizer, so the values are comparable. This is our measure of fluency.
\end{itemize}
Qwen3-8B was scored on the full MMLU set; because each question is scored on its own, taking the same subset from its saved per-question answers gives exactly the result of scoring only those questions.

\paragraph{Protocol.}
Every multiple-choice question is shown as plain text with the letters A to D, and the answer is the letter with the highest next-token probability.
No chat template is used for any model, so chat models and base models are scored the same way.
Most adapters were trained on 4-bit models (the books-only CPT adapter and the first 0.6B SFT stage used 16-bit models); all are applied to the 16-bit model they were trained from.
For every trained model we compare its answers with those of its starting model on the same questions, and report a paired 95\% bootstrap interval over questions.
Each model was trained once, so these intervals reflect test sampling only, not training variation.

\paragraph{Letter preference.}
Scoring by the most likely letter is known to be sensitive to a model's preference for particular letters \citep{zheng2023selectors}.
The correct answers in Belebele are spread close to evenly over the four letters (at most \BelGoldTop\% on one letter), so for each model we also report how often it chose its most frequent letter (Table~\ref{tab:bias}).
The 8B models choose their most frequent letter on at most \EightTaxBelTop\% of the Belebele questions, but stock Qwen3-0.6B chooses ``\SixStockBelTopL'' on \SixStockBelTop\% of the Belebele questions, and the June 0.6B adapter chooses ``\SixLightEnMTopL'' on \SixLightEnMTop\% of the English MMLU questions.
For such models a change in accuracy mostly reflects a change in letter preference, and we do not read it as a change in understanding.
We stored only the chosen letter, not the four probabilities, so we cannot correct for this preference after the fact.

\section{Results}
\label{sec:results}

\begin{figure}[t]
\centering
\includegraphics[width=\columnwidth]{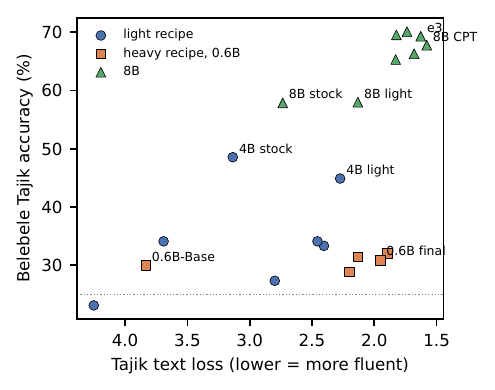}
\caption{Tajik text loss (lower is better, axis reversed) against Belebele Tajik accuracy. The dotted line is chance. Every trained model has a lower text loss than its starting model; Belebele rose clearly only for the 8B models with continued pretraining. The 0.6B scores near chance mostly reflect a preference for one answer letter (Table~\ref{tab:bias}).}
\label{fig:fluency}
\end{figure}

Table~\ref{tab:main} gives all scores, Figure~\ref{fig:fluency} plots text loss against Belebele accuracy, and Table~\ref{tab:bias} shows how strongly each model prefers one answer letter.

\begin{table*}[t]
\centering\small
\begin{tabular}{lrrr}
\toprule
Model & Text loss $\downarrow$ & Belebele Tajik & English MMLU \\
 & (\NPassages{} passages) & (\NBel) & (\NEnM) \\
\midrule
\multicolumn{4}{l}{\textit{Light recipe, June 2025 (attention-only LoRA, at most about \ATokensM M tokens per epoch)}} \\
Qwen3-0.6B & 4.25 & 23.1 & 41.5 \\
\quad + light adapter (QwenTg-v5) & 2.80 ($-$1.45) & 27.3 (+4.2$^{*}$) & 32.1 ($-$9.4$^{*}$) \\
Qwen3-1.7B & 3.69 & 34.1 & 58.0 \\
\quad + light adapter v1 & 2.40 ($-$1.29) & 33.3 ($-$0.8) & 54.7 ($-$3.3$^{*}$) \\
\quad + light adapter v2 & 2.46 ($-$1.23) & 34.1 (+0.0) & 54.3 ($-$3.7$^{*}$) \\
Qwen3-4B & 3.14 & 48.6 & 68.3 \\
\quad + light adapter & 2.27 ($-$0.86) & 44.9 ($-$3.7$^{*}$) & 67.9 ($-$0.4) \\
\midrule
\multicolumn{4}{l}{\textit{Heavy recipe at 0.6B, September to October 2025}} \\
Qwen3-0.6B-Base & 3.83 & 30.0 & 48.8 \\
\quad + SFT, 3 epochs & 2.13 ($-$1.70) & 31.4 (+1.4) & 42.6 ($-$6.2$^{*}$) \\
\quad + SFT, 8 epochs & 2.20 ($-$1.63) & 28.9 ($-$1.1) & 40.6 ($-$8.2$^{*}$) \\
\quad + CPT, 2 epochs & 1.90 ($-$1.94) & 32.1 (+2.1) & 40.3 ($-$8.5$^{*}$) \\
\quad + SFT, 2 epochs (final) & 1.95 ($-$1.88) & 30.9 (+0.9) & 41.2 ($-$7.6$^{*}$) \\
\midrule
\multicolumn{4}{l}{\textit{8B, June to October 2025}} \\
Qwen3-8B & 2.74 & 57.9 & 73.5 \\
\quad + light adapter (June) & 2.13 ($-$0.60) & 58.0 (+0.1) & 74.0 (+0.5) \\
\quad + CPT, books only & 1.83 ($-$0.91) & 65.3 (+7.4$^{*}$) & 75.4 (+1.9$^{*}$) \\
\quad + CPT, books, wiki, news & 1.58 ($-$1.16) & 67.8 (+9.9$^{*}$) & 73.2 ($-$0.3) \\
\quad + CPT + SFT (e3) & 1.63 ($-$1.11) & 69.3 (+11.4$^{*}$) & 72.3 ($-$1.2) \\
\quad + CPT + SFT + maths (e3.1) & 1.82 ($-$0.91) & 69.6 (+11.7$^{*}$) & 69.6 ($-$3.9$^{*}$) \\
\quad + CPT + SFT + maths, merged file & 1.82 ($-$0.91) & 69.6 (+11.7$^{*}$) & 69.6 ($-$3.9$^{*}$) \\
\quad + CPT + SFT + maths + exam and Tajikistan & 1.74 ($-$1.00) & 70.1 (+12.2$^{*}$) & 70.8 ($-$2.7$^{*}$) \\
\quad + CPT + SFT + tax code & 1.68 ($-$1.05) & 66.3 (+8.4$^{*}$) & 71.4 ($-$2.1) \\
\bottomrule
\end{tabular}
\caption{Results for all \NModels{} rows (the two \EMath{} rows are the same model, as an adapter and as a merged file). Text loss is the mean negative log-likelihood per token on the Tajik Belebele passages (lower means the model finds real Tajik text more predictable). The other two columns are accuracy (\%) on four-option questions, where guessing scores 25. English MMLU uses \NEnM{} questions (every \MmluStride th item). In brackets: change against the stock model at the top of each block (the unindented row), on the same questions; $^{*}$ marks changes whose paired 95\% bootstrap interval excludes zero.}
\label{tab:main}
\end{table*}

\begin{table*}[t]
\centering\small
\begin{tabular}{lrr}
\toprule
Model & Belebele Tajik & English MMLU \\
\midrule
Qwen3-0.6B & 95 & 38 \\
\quad + light adapter (QwenTg-v5) & 48 & 45 \\
Qwen3-1.7B & 37 & 35 \\
\quad + light adapter v1 & 46 & 32 \\
\quad + light adapter v2 & 49 & 32 \\
Qwen3-4B & 36 & 28 \\
\quad + light adapter & 37 & 30 \\
Qwen3-0.6B-Base & 48 & 35 \\
\quad + SFT, 3 epochs & 43 & 36 \\
\quad + SFT, 8 epochs & 53 & 34 \\
\quad + CPT, 2 epochs & 56 & 57 \\
\quad + SFT, 2 epochs (final) & 55 & 52 \\
Qwen3-8B & 33 & 29 \\
\quad + light adapter (June) & 29 & 27 \\
\quad + CPT, books only & 31 & 30 \\
\quad + CPT, books, wiki, news & 36 & 30 \\
\quad + CPT + SFT (e3) & 31 & 29 \\
\quad + CPT + SFT + maths (e3.1) & 32 & 29 \\
\quad + CPT + SFT + maths + exam and Tajikistan & 31 & 28 \\
\quad + CPT + SFT + tax code & 40 & 30 \\
\bottomrule
\end{tabular}
\caption{Share (\%) of questions on which each model chose its most frequent letter. The correct answers are close to evenly spread (at most \BelGoldTop\% on one letter in Belebele), so a share far above a quarter means the model is mostly choosing by position.}
\label{tab:bias}
\end{table*}

\subsection{Light adapters: lower text loss, no better comprehension}

Every light adapter lowered text loss: by \SixLightDTlAbs{} at 0.6B, \OneSevVOneDTlAbs{} at 1.7B, \FourLightDTlAbs{} at 4B and \EightLightDTlAbs{} at 8B.
None improved Belebele at 1.7B, 4B or 8B.
At 1.7B the two versions changed it by \OneSevVOneDBel{} and \OneSevVTwoDBel{} points, at 8B by \EightLightDBel{} (interval \EightLightDBelLo{} to \EightLightDBelHi), and at 4B it fell by \FourLightDBelAbs{} points (interval \FourLightDBelLo{} to \FourLightDBelHi).
At 0.6B the adapter raised Belebele by \SixLightDBel{} points (interval \SixLightDBelLo{} to \SixLightDBelHi), but both models are near chance and the stock model chose one letter on \SixStockBelTop\% of the questions against \SixLightBelTop\% after training (Table~\ref{tab:bias}), so the change is a change in letter preference, not evidence of understanding.
English MMLU fell by \OneSevVOneDEnMAbs{} and \OneSevVTwoDEnMAbs{} points for the two 1.7B versions and by \SixLightDEnMAbs{} at 0.6B, where the letter preference also changed.

In 2025 we judged these models by whether they wrote Tajik, and they did.
The scores show that writing Tajik and understanding it are different things, and the light recipe changed only the first.

\subsection{8B: comprehension rose with continued pretraining}

After CPT on books alone, Belebele was \EightBooksDBelAbs{} points higher than for Qwen3-8B (interval \EightBooksDBelLo{} to \EightBooksDBelHi), and after CPT on the full corpus \EightCptDBelAbs{} points higher (interval \EightCptDBelLo{} to \EightCptDBelHi).
With SFT on top, \ESft{} scores \EightSftBel{} on Belebele against \EightStockBel{} for Qwen3-8B, with an interval above zero.
Its English MMLU changed by \EightSftDEnM{} points (interval \EightSftDEnMLo{} to \EightSftDEnMHi): a loss of more than \EightSftDEnMLoAbs{} points is excluded, a smaller one is not.

The June light adapter and the CPT adapters start from the same model, but they differ in training text, adapter size and precision, and the June adapter's data was not recorded.
We therefore cannot say which of these differences made the CPT models better; we can only say that comprehension rose in the runs with CPT and not in the run without it.

\subsection{0.6B: lower text loss, answers still near chance}

Qwen3-0.6B-Base with the full heavy recipe reached a text loss of \SixFinalTl, close to the \EightBooksTl{} of the 8B model after books-only CPT and above the \EightCptTl{} after CPT on the full corpus.
Its Belebele accuracy stayed near chance (\SixFinalBel), and it chose its most frequent letter on \SixFinalBelTop\% of the Belebele questions.
English MMLU fell by \SixSftADEnMAbs{} to \SixCptDEnMAbs{} points at every stage, and the letter preference on Belebele grew at the CPT and final stages.
We see no evidence that the 0.6B model understands Tajik questions.
Its answers in 2025, which we did not score, pointed the same way: fluent Tajik with wrong facts, such as naming Dushanbe the capital of Japan.

\subsection{Later 8B branches}

The maths model (\EMath), trained on top of \ESft{} with a file named as exam test data whose contents were never documented, changes Belebele by \EightMathVsSftBel{} points against \ESft{} and lowers English MMLU by \EightMathVsSftEnMAbs{} points.
The MultiStage model, trained on top of \EMath{} with entrance-exam and Tajikistan questions, changes Belebele by \EightMultiVsSftBel{} points against \ESft{} and English MMLU by \EightMultiVsSftEnM.
Neither branch improves reading comprehension beyond \ESft{} by more than the noise, and both have a lower English MMLU point estimate than \ESft{} (we did not compute paired intervals against \ESft), so we do not count them as progress.
The tax-code branch, which saw neither the maths nor the exam data, is \EightTaxVsSftBelAbs{} points below \ESft{} on Belebele and \EightTaxVsSftEnMAbs{} points below it on English MMLU.

\section{What Went Wrong}
\label{sec:failures}

We report these because a small lab is likely to repeat them.

\paragraph{No held-out test in 2025.}
Every 2025 model was judged by reading its generated answers.
Those answers showed Tajik letters and Tajik words, which every model produced, so the light adapters looked as successful as the 8B model.
The tax-code data had \NTaxHeldDeleted{} questions set aside for testing, but they were deleted before training and never used.

\paragraph{Lost training records.}
CPT epochs one to three of the 8B model have no loss record; only the fourth epoch survives, and its loss is flat (\BwnLossFirst{} to \BwnLossLast) because the adapter was already trained.
The training runs behind the June 8B adapter, the books-only CPT adapter and the merged \EMath{} model left no notebook, and the June 0.6B adapter left no record of its data.
We reconstructed the history in this report from notebooks, Hugging Face upload times and adapter files.

\paragraph{A thinking-mode dataset without thinking.}
The 8B SFT data was built to teach step-by-step reasoning, but only \ThmReasonPct\% of it contains reasoning.
All \NThmEval{} answers we generated from the trained models start with an empty reasoning block. This is weak evidence on its own, since the chat template can insert such a block, but it agrees with the data.

\paragraph{A 4-bit release without layers.}
In October 2025 we quantised \ESft{} with SINQ and A-SINQ \citep{muller2025sinq} and uploaded the results.
Each upload holds \SinqTensors{} tensors (\SinqGb{} GB): the input and output embeddings and the normalisation weights.
The transformer layers are missing; the most likely cause is that the quantised model was saved with the standard save call of the training library, which does not store the quantised layer type.
The size reduction noted at the time measured the embeddings alone, and no quality test was run.
These files cannot be evaluated, and we do not report a 4-bit result.

\section{Discussion}
\label{sec:discussion}

\paragraph{Why text loss fell everywhere.}
Lowering text loss on Tajik requires the model to predict Tajik word forms and common phrases, which a small adapter can learn from tens of millions of tokens.
Answering a question about a Tajik passage also requires the model to connect the Tajik words to what it already knows.
Our results suggest that the light adapters taught the first and not the second, and at the smaller sizes they cost English knowledge as well.

\paragraph{Size and recipe.}
The same CPT corpus and similar SFT data were followed by better comprehension at 8B and not at 0.6B.
Qwen3-0.6B-Base scores \SixBaseEnM{} on English MMLU against \EightStockEnM{} for Qwen3-8B, so it has far less knowledge to carry over into Tajik, and our training reduced its English score further.
We read this as a hypothesis, not a finding: the comparison has only two sizes for the heavy recipe and is not controlled (Table~\ref{tab:cells}).
The 0.6B run started from the base model rather than the chat model, trained on answers only, and used a larger adapter, and the 1.7B and 4B models never received the heavy recipe.

\paragraph{What we would do differently.}
\begin{itemize}
\item Hold out a test set before training, and score every checkpoint on it, not only on generated samples.
\item Measure comprehension and text loss separately; a falling text loss says nothing about reading comprehension.
\item Store the probability of every answer option, so that a preference for one letter can be measured and corrected.
\item Check that a quantised file loads and scores before releasing it.
\end{itemize}

\section{Release}
\label{sec:release}

The starting models are the Qwen3 releases. All trained models in Table~\ref{tab:main} are on Hugging Face under \texttt{SaidzodaEng}, except the June 8B adapter, which is private.
The main 8B model is \RepoSft{} and the final 0.6B adapter is \RepoSixFinal.
Some repository names do not match their contents; for example, the repository named as the merged MultiStage model holds the same weights as \ESft, and the quantised repositories hold no layers (Section~\ref{sec:failures}).
The evaluation script, the per-question answers of every model on the three tests of this report, and the scripts that build every number, table and figure in this report are available at \RepoResults.

\section{Conclusion}
\label{sec:conclusion}

We scored the Qwen3 models we adapted to Tajik between June and October 2025.
Small adapters lowered Tajik text loss at every size and improved reading comprehension at no size where the score is above chance.
At 8B, Belebele rose by \EightCptDBelAbs{} points in the run with continued pretraining on about \CptTokensM{} million tokens, and our main model, \ESft, is \EightSftDBelAbs{} points above Qwen3-8B on Belebele with no large loss in English.
At 0.6B the heavy recipe lowered text loss while the answers stayed near chance.
The lesson we take is to measure comprehension, not only fluency, from the first run.

\section*{Limitations}

\paragraph{Not a controlled comparison.}
The four groups differ in start model (chat model or base model), data, adapter size, precision and loss masking (Table~\ref{tab:cells}).
The training data of the June 8B and 0.6B adapters is unknown.
Our conclusions about size and recipe rest on consistent patterns, not on runs that change one factor at a time.

\paragraph{Single runs.}
Each model was trained once, and the intervals cover only the sampling of test questions.
Most adapters were trained on 4-bit models and all were evaluated on 16-bit ones.

\paragraph{Scoring.}
All models were scored by the most likely answer letter, without a chat template.
Understanding is measured by one test, Belebele Tajik, a translated reading task; we have no Tajik knowledge or generation test in this report.
Chat models may score differently in their own format, and the light adapters were trained inside a chat template, so part of any change may be a change in format.
The smallest models prefer one letter strongly (Table~\ref{tab:bias}); we kept only the chosen letter, so we cannot correct for this, and their accuracies near chance carry little information.
English MMLU uses \NMmluSubset{} of the \NMmluFull{} questions, and generation quality was not measured.

\paragraph{Possible overlap with tests.}
We did not check any training data against the test sets.
The CPT corpus includes Tajik Wikipedia and the SFT data includes question answering and Wikipedia rewritten as dialogue; these could overlap with Belebele passages, which would make the gains of \ESft{} and the text loss of the CPT models optimistic.
The Belebele passages come from FLORES news and travel text, which is not part of our corpora by design, but we did not run a check.

\paragraph{Reconstructed history.}
Several training records are lost (Section~\ref{sec:failures}).
Where a setting is not recorded, we say so rather than infer it.
Some repository names on Hugging Face do not match their contents (Section~\ref{sec:release}).

\section*{Use of AI Assistants}

We used AI assistants (Anthropic Claude models) to read our 2025 training notebooks and reconstruct the history of the runs, to write the evaluation and analysis scripts, to search for and check references, and to draft parts of the text.
The authors designed and ran the training in 2025, checked the reconstructed history and every number against the recorded files, and take full responsibility for the content.
Every number in this report is generated by a script from logged files, and each reference was checked against its arXiv, ACL Anthology or DOI record.
AI assistants are not authors.

\bibliography{refs}
\bibliographystyle{acl_natbib}

\end{document}